\pdfoutput=1

\documentclass[10pt,twocolumn,letterpaper]{article}

\usepackage{preprintstyle}

\usepackage{booktabs}
\usepackage{multirow}
\usepackage{graphicx}
\usepackage{xspace}
\usepackage{amsmath,amssymb}
\usepackage{colortbl}
\usepackage{pifont}

\usepackage{cuted}

\usepackage{currfile}

\usepackage{caption}
\newcommand{\yes}{\ding{51}}
\newcommand{\no}{\ding{55}}

\newcommand{\method}{SkimPrefill\xspace}

\definecolor{ourrow}{RGB}{229,241,233}

\definecolor{linkblue}{rgb}{0.21,0.49,0.74}
\usepackage[pagebackref,breaklinks,colorlinks,allcolors=linkblue]{hyperref}

\title{Let a Small Model Read First:\\ Faster Prefill for Long Interleaved Image-Text Inputs}

\author{
  \textbf{Xiaolin Lin\textsuperscript{1}},
  \textbf{Kai-Chiang Wu\textsuperscript{2}},
  \textbf{Bing Li\textsuperscript{3}},
  \textbf{Grace Li Zhang\textsuperscript{1}}
\\
\\
  \textsuperscript{1}Technische Universität Darmstadt,
  \textsuperscript{2}National Yang Ming Chiao Tung University,
  \textsuperscript{3}TU Ilmenau
\\
  \small{
    \texttt{xiaolin.lin@tu-darmstadt.de}, \texttt{kcw@cs.nycu.edu.tw},
  }
\\
  \small{
    \texttt{bing.li@tu-ilmenau.de}, \texttt{grace.zhang@tu-darmstadt.de}
  }
}

\begin{document}
\maketitle
\begin{abstract}
Vision-language models (VLMs) can answer questions about long interleaved inputs, such as documents of many pages and collections of hundreds of images, but they are slow to start: on inputs of about 100K tokens, the prefill takes most of the wait for the first output token.
A common approach prunes the input by scoring its tokens with the target model itself or with a small model of the same family that first writes an answer.
However, this scoring takes much of the time that pruning saves.
We find that the evidence for a question lies in a few whole units (images or paragraphs), and that a small hybrid model finds them with one attention head of an early layer, even at half image resolution.
Based on these findings, we propose \method, which lets the small hybrid model read the whole input first, up to that early layer, with images at half resolution.
Most layers of the small hybrid model (Qwen3.5-0.8B) run in linear time, so this pass is short.
\method then ranks every unit with that head and gives the target the top units as an ordinary shorter prompt. One untrained selector serves targets of different model families.
Extensive experiments on MMLongBench at 128K tokens, with 19 datasets of five task types, show the effectiveness of \method.
With 10\% of the input tokens kept, \method speeds up the prefill of Qwen3.5-9B, Gemma-4-12B and Qwen3-VL-32B by 3.7$\times$, 11.6$\times$ and 8.2$\times$ while maintaining their average accuracy.
\end{abstract}

\begin{figure}[t]
\centering
\includegraphics[width=\linewidth]{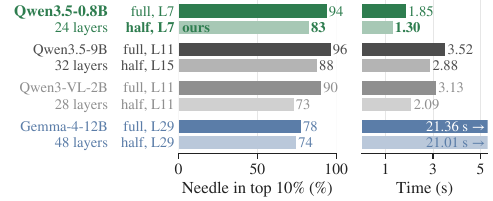}
\caption{How well and how fast a model finds the evidence, on 196 MM-NIAH questions outside the test set. One head of one layer scores the images of a question. \emph{Needle in top 10\%}: at a budget of about 10\% of the images, how often the images ranked highest by the head include a needle, which is an image holding the answer. We show the best head of the best layer. \emph{Time}: running the model up to that layer, preprocessing included. Full and half: image resolution; L7: layer 7.}
\label{fig:finding}
\end{figure}

\begin{figure*}[t]
\centering
\includegraphics[width=\textwidth]{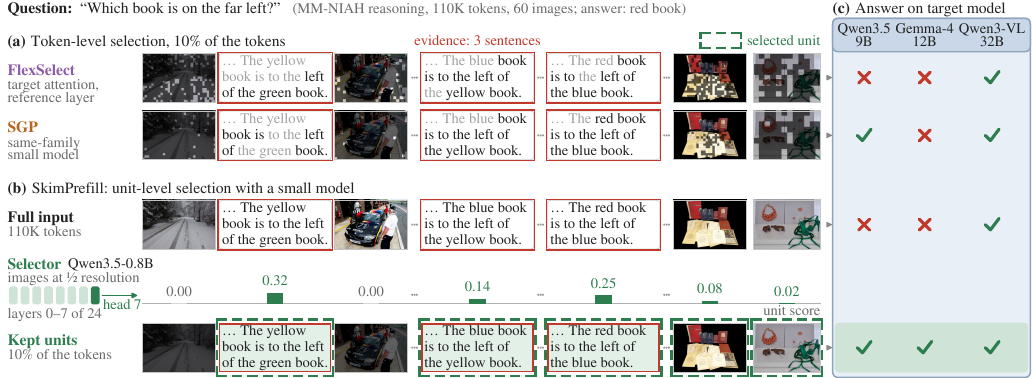}
\caption{A needle-in-a-haystack question from MMLongBench, with seven of its 1,197 images and paragraphs in input order; the evidence is three sentences.
(a) FlexSelect and SGP keep 10\% of the tokens (dropped patches darkened, dropped words faded; their Qwen3.5-9B masks); both drop ``yellow'' and ``blue'' from the evidence.
(b) \method: Qwen3.5-0.8B reads the whole input, one head of layer 7 scores each image and paragraph, and every target reads the same kept units (green dashed frames), which hold as many tokens as FlexSelect and SGP keep.
(c) Whether each target answers the question based on the kept tokens, the full input or the kept units, respectively.}
\label{fig:teaser}
\end{figure*}

\section{Introduction}
\label{sec:intro}
Vision-language models (VLMs)~\cite{bai2025qwen3vl,google2026gemma4,qwen2026qwen35} answer questions about long interleaved inputs, such as documents of many pages~\cite{ma2024mmlongbench} and collections of hundreds of images~\cite{wang2025mmlongbench,wu2025visual}, but respond slowly.
A key source of this delay is the prefill~\cite{li2025mminference}, which passes every token of these inputs, up to 128K~\cite{wang2025mmlongbench}, through every layer before the first output token.
However, a question usually depends on a small part of such an input.
Token pruning~\cite{chen2024image,zhang2025sparsevlm,yang2025visionzip,zhao2025stitch} therefore shortens the target's prefill by dropping tokens that are redundant or that the answer does not need.

Most existing VLM pruning methods score the input with the target or with a small model of its family, but they keep individual or merged tokens~\cite{yang2025visionzip,zhang2025sparsevlm} instead of a whole image or a whole paragraph (\cref{tab:related}).
FastV~\cite{chen2024image} ranks visual tokens by the attention of an early layer of the target and prunes them inside it.
This is cheap, but one early layer ranks tokens poorly at small budgets~\cite{zhao2025stitch}.
The training-free form of FlexSelect~\cite{zhang2025flexselect} takes the attention of a deeper reference layer instead, so the target reads the whole input up to that layer before it selects.
Small-VLM-Guided Pruning (SGP)~\cite{zhao2025stitch} lets a small model of the target's family write an answer first and sums its attention over all layers.
IF-Prune~\cite{sun2026ifprune} moves beyond this answer-driven attention: it fine-tunes a small model that scores each visual token in one pass and serves larger models of the same architecture.
SGP and IF-Prune pass token positions to the target, so their small models serve only targets of their own family.

\begin{table*}[t]
\centering
\caption{Methods that shorten a long input before or inside the target. Other families: the kept input can go to a target of another family.}
\label{tab:related}
\footnotesize
\setlength{\tabcolsep}{5pt}
\begin{tabular}{lcccll}
\toprule
Method & Training-free & VLM & Other families & Kept unit & Scorer \\
\midrule
FastV~\cite{chen2024image} & \yes & \yes & \no & image tokens & target, early layer \\
FlexSelect~\cite{zhang2025flexselect} (training-free form) & \yes & \yes & \no & video tokens & target, middle layer \\
SGP~\cite{zhao2025stitch} & \yes & \yes & \no & image tokens & small VLM, all layers \\
IF-Prune~\cite{sun2026ifprune} & \no & \yes & \no & image tokens & trained small VLM \\
LongLLMLingua~\cite{jiang2024longllmlingua} & \yes & \no & \yes & documents, tokens & small LM \\
SpecPrefill~\cite{liu2025speculative} & \yes & \no & \no & text blocks & small LLM, all layers \\
Cross-Family SpecPrefill~\cite{upasani2026cross} & \yes & \no & \yes & text blocks & small LLM, all layers \\
CAPS~\cite{zhu2026attention} & \no & \yes & \yes & pages & small VLM, one head \\
\rowcolor{ourrow} \method & \yes & \yes & \yes & images, paragraphs & small VLM, early head \\
\bottomrule
\end{tabular}
\end{table*}

Keeping individual tokens in the state-of-the-art work breaks the evidence apart on long interleaved inputs. In \cref{fig:teaser}a, the kept tokens spread over most images and paragraphs, and words of the evidence are dropped.
On such fragments, the target also often thinks longer before it answers, which adds to the waiting time.
A good selector for such inputs should therefore be fast and accurate: it should cost much less than the prefill and keep the evidence whole.

A central observation behind our method is that a small model can find the evidence that a large one needs.
This evidence lies in a few whole images or paragraphs, and an early layer of the small model already points to them.
One attention head of that layer ranks the needle image near the top almost as often as the best layer of a much larger model of the same family (\cref{fig:finding}).
It does so even more often than the best layers of larger models from other families.
Motivated by this, we propose \method (\cref{fig:teaser}b), which lets the small hybrid model read the whole input first and stop at that layer.
The small model, which we call the selector, then scores every image and paragraph with that head and passes the top units to the target as an ordinary, shorter prompt.

The selector is fast because it is small and stops at that early layer, and accurate because it keeps whole images and paragraphs (\cref{fig:units}), ranked by that head.
It runs only the first 8 of the 24 layers of Qwen3.5-0.8B~\cite{qwen2026qwen35}, most of which use linear attention, on images at half resolution.
On Qwen3-VL-32B, the selection takes 0.71\,s, while the target needs 17.58\,s to prefill the full input.
Its ranking of whole units keeps most of the evidence in the input (\cref{fig:evidence}), and every target then thinks less than with the full input.

With \method, the target also runs in an unchanged serving engine such as SGLang~\cite{zheng2023sglang}, since it receives an ordinary, shorter prompt.
FastV and SGP drop tokens inside the target and FlexSelect reads the attention of its reference layer, so each target model needs its own engine changes.
\method needs them once, for its selector, which serves targets of other families as well.
Unlike IF-Prune~\cite{sun2026ifprune}, our selector needs no training.
It needs only 50 synthetic documents, which share no text or image with the evaluation data, to find its head and a threshold.
Here we summarize our contributions:
\begin{itemize}
\item We find that, on long interleaved inputs, keeping whole images and paragraphs instead of single tokens raises the accuracy of three different scorers (\cref{sec:exp_ablation}). Single tokens break the evidence apart, and the target then thinks about as long as with the full input or longer (\cref{fig:units}).
\item We show that one attention head in an early layer of a small model finds the evidence almost as often as the best layer of a much larger model.
\item We propose \method, which gives the target the whole images and paragraphs that the head above ranks highest as an ordinary, shorter prompt, so one selector serves targets of different families without training.
\item Extensive experiments on MMLongBench~\cite{wang2025mmlongbench} at 128K tokens, with 19 datasets of five task types, show the effectiveness of \method.
With 10\% of the input tokens kept, it speeds up the prefill of three target models, Qwen3.5-9B, Gemma-4-12B and Qwen3-VL-32B, by 3.7$\times$, 11.6$\times$ and 8.2$\times$ while maintaining their average accuracy (\cref{tab:main}).
\end{itemize}

\section{Related work}
\label{sec:related}
\Cref{tab:related} compares the pruning methods closest to ours; \method is the one that serves VLM targets of other families without training.

\paragraph{Selecting tokens with the target.}
Most visual token reduction methods let the target decide what to drop.
FastV~\cite{chen2024image}, SparseVLM~\cite{zhang2025sparsevlm} and PyramidDrop~\cite{xing2025pyramiddrop} rank visual tokens by their attention at one or several layers of the target, and VisionZip~\cite{yang2025visionzip} keeps the dominant tokens of the vision encoder and merges the rest.
GemFilter~\cite{shi2024discovering} selects text tokens with the early layers of an LLM, and FlexSelect~\cite{zhang2025flexselect} ranks video tokens with one reference layer of the target, or with a small selector trained to mimic it.
In all of them the vision encoder or the early layers of the target read the whole input.
Sparse-attention methods such as MInference~\cite{jiang2024minference} and MMInference~\cite{li2025mminference} keep every token and speed up only the attention computation of the prefill.
KV-cache compression such as LOOK-M~\cite{wan2024lookm} drops or merges cached image tokens during the prefill, which saves memory and decoding time but not prefill time.

\paragraph{Selecting tokens with a small model.}
SGP~\cite{zhao2025stitch} and IF-Prune~\cite{sun2026ifprune} select with a small model of the target's family or architecture, LAST~\cite{yang2026last} scores visual tokens with a compact VLM, and VisCache~\cite{wang2026viscache} filters video keyframes with CLIP.
For text-only LLMs, LLMLingua~\cite{jiang2023llmlingua} drops whole in-context examples and then tokens by the perplexity of a small model, LongLLMLingua~\cite{jiang2024longllmlingua} first ranks documents by the perplexity of the question given each document, and LLMLingua-2~\cite{pan2024llmlingua2} trains a token classifier.
SpecPrefill~\cite{liu2025speculative} keeps fixed-size blocks scored by every layer and head of a draft LLM, which can come from another family if the kept text is re-tokenized~\cite{upasani2026cross}.
Coarser units are also used in text: ChunkKV~\cite{liu2025chunkkv} keeps or drops the target's cached tokens in chunks and CPC~\cite{liskavets2025cpc} removes whole sentences, since removing single tokens can break the context.
\method selects whole images and paragraphs of an interleaved input for VLM targets, and ranks images and text in one pass with one head of an early layer.
With the same scores, these units give higher accuracy than single tokens and shorter reasoning (\cref{sec:exp_ablation}).

\paragraph{Retrieval heads in LLMs and VLMs.}
In LLMs, a fraction of the heads perform long-context retrieval~\cite{wu2024retrieval,xiao2024duoattention}, and CompressKV~\cite{lin2025compresskv} keeps the KV-cache entries they attend to.
ICR~\cite{chen2025attention} and QRHead~\cite{zhang2025qrhead} re-rank passages by the query's attention to them in an LLM, over all heads or a few, and in VLMs such heads attend to visual evidence and rank pages and regions~\cite{li2026retrieval,pei2026vera}.
CAPS~\cite{zhu2026attention} fine-tunes a VLM and selects document pages with its best head, scoring each page on its own.
\method uses one such head of a small model, without training, to choose another model's input.

\paragraph{Retrieval-augmented generation.}
Vision-based multimodal RAG keeps whole pages or images chosen by a retriever that encodes each on its own, with one embedding per unit (GME~\cite{zhang2025gme}) or one per patch (ColPali~\cite{faysse2025colpali}, ColEmbed V2~\cite{moreira2026colembed}).
M3DocRAG~\cite{cho2024m3docrag} uses ColPali, VisRAG~\cite{yu2024visrag} its own VLM-based retriever, and SV-RAG~\cite{chen2025svrag} an adapter of the answering VLM.
In contrast, the selector of \method reads all units together in one pass, with a model not trained for retrieval.
\Cref{sec:exp_ablation} compares it with GME, ColEmbed V2 and the image-text encoder SigLIP 2~\cite{tschannen2025siglip2}.

\section{What a fast and accurate selector needs}
\label{sec:analysis}
This section measures what a fast and accurate selector needs on interleaved inputs of about 100K tokens.
We find that most of the wait for the first token is the target's prefill, the evidence for a question lies in a few whole units, and one early head of a small hybrid model finds these units at a fraction of the prefill's cost.

\paragraph{Most of the wait for the first token is the target's prefill.}
In \cref{fig:speed}, an input of about 105K tokens on Qwen3-VL-32B waits 19.06\,s for the first output token, of which 17.58\,s is the target's prefill.
On Qwen3.5-9B, the prefill takes 5.82 of the 7.19\,s (see \cref{sec:suppl_speed}).
A selector that shortens the target's input can remove most of this wait, as long as choosing what to keep costs much less than the prefill itself.

\begin{figure}[t]
\centering
\includegraphics[width=\linewidth]{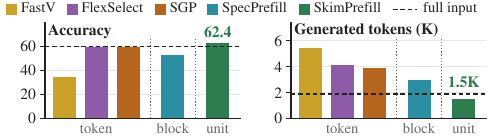}
\caption{Accuracy and generated tokens per answer on Qwen3-VL-32B with 10\% of the input tokens kept.
FastV, FlexSelect and SGP select at the token level, SpecPrefill at the block level (32 consecutive tokens) and \method at the unit level (images and paragraphs); dashed: the full input.}
\label{fig:units}
\end{figure}

\paragraph{The evidence lies in a few whole units.}
The evidence for a question usually comes in whole units: an image, a page or a paragraph.
For example, to count objects or read a table, the target needs the whole image or page.
The data files of MMLongBench~\cite{wang2025mmlongbench} mark the evidence of 928 of our test questions: needle images, needle sentences or answer pages.
This evidence is small: on average, up to 2.4\% of the input tokens, depending on the task.
Even kept as whole units, it takes up at most 10\% of the tokens for 922 of these questions.
\Cref{fig:units} compares token-, block- and unit-level selection on Qwen3-VL-32B with 10\% of the tokens.
Keeping the most important tokens also works well: the accuracy degradation of FlexSelect and SGP relative to the full input is up to 0.6 points.
However, the tokens that token-level methods~\cite{chen2024image,zhang2025flexselect,zhao2025stitch} consider important are spread over most of these units and miss words of the evidence (\cref{fig:teaser}a), and the target then generates about as many tokens per answer as with the full input or more.
SpecPrefill's blocks of 32 tokens also lengthen the answers (2.9K against 1.9K tokens) and lower the accuracy by 6.5 points.
Keeping whole units instead makes the target generate fewer tokens than with the full input and answer more accurately.
A selector should therefore rank and keep whole units.

\paragraph{An early head of a small hybrid model finds these units.}
We find that an early head of a small hybrid model locates these units about as well as larger models, or better, and in less time (\cref{fig:finding}).
On the image questions of MM-NIAH, a needle-in-a-haystack benchmark, head 7 of layer 7 of Qwen3.5-0.8B puts a needle among the top-ranked images for 94\% of the questions, against 96\% for the best layer of Qwen3.5-9B, 90\% for Qwen3-VL-2B and 78\% for the larger Gemma-4-12B.
It reaches that layer in 1.85\,s, against 3.52, 3.13 and 21.36\,s for these models.
At half resolution, which \method uses, the rate is still 83\%, in 1.30\,s.
Earlier work shows that hybrid models with few attention layers do well on needle-in-a-haystack tests of long inputs~\cite{lieber2024jamba,waleffe2024empirical}.
Layers with a fixed-size state retrieve exact content from the input less well than attention~\cite{jelassi2024repeat,yang2025gated}, and hybrid models depend on their attention layers for retrieval~\cite{bick2025understanding}.
\method picks its head on synthetic documents, as described in \cref{sec:skim}.

\section{Method}
\label{sec:method}

\method lets a small hybrid model read the whole input before the target does.
The selector runs up to an early full-attention layer, scores every image and paragraph with one head there, and passes the highest-scoring units, as a whole, to the target as an ordinary, shorter prompt.

\subsection{Preliminaries}
\label{sec:prelim}
A VLM, the target, answers a question about a long document of interleaved images and text.
We write the input as
\begin{equation}
X=[x_{\text{inst}},\,u_1,\dots,u_N,\,x_q],
\end{equation}
where $x_{\text{inst}}$ is the instruction, $x_q$ the question, and $u_1,\dots,u_N$ the units of the document in reading order.
A unit is a paragraph of text, or an image together with any short label next to it in the input, such as ``Document gao-12-156 (page 24):'' before the image or ``label: 10'' after it.
Given a budget ratio $\rho$, we look for a set $S$ of units such that the shorter input
\begin{equation}
X_S=[x_{\text{inst}},\,(u_i)_{i\in S},\,x_q]
\end{equation}
holds about $\rho$ of the tokens of $X$, and the target answers as accurately based on it as based on $X$.
The budget is counted in tokens, but each model turns the same text and images into a different number of tokens.
We therefore count them once for all targets, with the tokenizer and image processor of the selector and with every image at its original resolution, which is how the target reads the kept images.
The same units are then kept for every target.

\subsection{Selecting the input with a small hybrid model}
\label{sec:skim}

The selector has to rank the units at a small fraction of the prefill's cost.
Following \cref{sec:analysis}, we use the small hybrid model Qwen3.5-0.8B as the selector.
It reads the whole input $X$ once, with every image halved in width and height, and stops after layer $\ell$, its second full-attention layer (layer 7), where one head already finds the evidence; six of the eight layers it runs are Gated DeltaNet layers whose time grows linearly with the input length.

\paragraph{Scoring units with one head.}
At layer $\ell$, the question tokens $Q$ (the last tokens of the prompt and the tokens of any image in the question) attend to the whole document.
The score of unit $u_i$ is the attention that $Q$ pays to it in one head $h$, averaged over the unit's tokens so that a large image does not outrank a small one by its size alone:
\begin{equation}
s_i=\frac{1}{|u_i|}\sum_{q\in Q}\sum_{t\in u_i}A^{(\ell,h)}_{q\to t},
\end{equation}
where $A^{(\ell,h)}_{q\to t}$ is the attention from $q$ to $t$ in head $h$ of layer $\ell$, and $t$ runs over the visual tokens of an image unit.

\paragraph{Finding the retrieval head.}
Following retrieval-head studies~\cite{wu2024retrieval,lin2025compresskv,li2026retrieval}, we find this head with needle-in-a-haystack questions on synthetic documents that share no text or image with MMLongBench.
Each of the 50 documents mixes paragraphs of PG-19 books~\cite{rae2020compressive} with 64 pictures from MileBench~\cite{song2024milebench} and comes with one question about an image needle, a book cover pasted into or added among the pictures, and one about text needles in its paragraphs.
We pick the layer $\ell$ and the head $h$ whose top-ranked units contain the image needles most often over all these questions; our code release will include the scripts and the list of sources that rebuild the documents.

\subsection{Passing whole units to targets of different \texorpdfstring{\mbox{families}}{families}}
\label{sec:build}
With the units ranked, the selector fills the budget with whole units and passes them to the target as an ordinary, shorter prompt: the kept units stay in their original order, with their images at full resolution, and form $X_S$ of \cref{sec:prelim} with consecutive positions.
The target needs no access to the selector, to its attention or to its KV cache, so targets of different families read $X_S$ in an unchanged serving engine.

\paragraph{Fixed parts.}
The instruction, the question and the last $W$ tokens of the document are always kept, since the question's attention to the text right before it is high whatever that text holds.
The rest of the budget, $B$, goes to the units.

\paragraph{Splitting the budget across modalities.}
Images and text share one budget, but their scores are not on the same scale: an image can hold the evidence while it scores below the best text paragraph.
We therefore rank the units of each modality separately and first decide which modality is more likely to hold the evidence, the primary modality
\begin{equation}
m^*=\begin{cases}\text{image} & \text{if } \Delta\ge\tau,\\ \text{text} & \text{otherwise,}\end{cases}\quad \Delta=\log_{10}\frac{\max_{i\in\mathcal{I}}s_i}{\max_{i\in\mathcal{T}}s_i}
\label{eq:primary}
\end{equation}
where $\mathcal{I}$ and $\mathcal{T}$ are the image and the text units, and text units within the last $W$ tokens are left out of the maximum over $\mathcal{T}$.
Here $\tau$ sets how far the best image score may fall below the best text score before text becomes primary.
We set it on the synthetic documents of \cref{sec:skim}, to the value that best separates image questions from text questions.
The primary modality receives a fraction $\alpha$ of $B$ and the other modality the rest.
Each modality takes the units that still fit its budget in decreasing order of $s_i$, and the budget that one modality does not use goes to the other.

\section{Experiments}
\label{sec:experiments}

\begin{table*}[t]
\centering
\caption{Accuracy on MMLongBench at 128K tokens and prefill speedup, with 10\% of the input tokens kept.
\emph{Kept}: fraction of the input that the target prefills.
\emph{Speedup}: time to first token of the full input over that of the method, selector included.
\textbf{Bold}: best pruning method.}
\label{tab:main}
\small
\setlength{\tabcolsep}{5.1pt}
\begin{tabular}{cll*{6}{c}>{\hspace{3pt}}cc}
\toprule
& & & \multicolumn{6}{c}{Accuracy on MMLongBench, 128K} & \multicolumn{2}{c}{Prefill} \\
\cmidrule(lr){4-9}\cmidrule(l){10-11}
Target & Method & Selection by & VRAG & NIAH & ICL & Summ & DocQA & Avg. & Kept & Speedup \\
\midrule
 & Full input & -- & 62.0 & 60.5 & 85.5 & 22.5 & 72.1 & 60.5 & 100\% & \phantom{0}1.0$\times$ \\
 & FastV~\cite{chen2024image} & target, one layer & 61.0 & 42.0 & 50.8 & 18.4 & 21.0 & 38.6 & \phantom{0}10\% & \phantom{0}1.6$\times$ \\
 & FlexSelect~\cite{zhang2025flexselect} & target, reference layer & 62.0 & 64.7 & 85.5 & 20.6 & \textbf{75.0} & 61.6 & \phantom{0}10\% & \phantom{0}0.9$\times$ \\
 & SGP~\cite{zhao2025stitch} & small model, same family & 60.5 & 59.7 & 83.8 & \textbf{21.0} & 73.2 & 59.6 & \phantom{0}10\% & \phantom{0}1.1$\times$ \\
 & SpecPrefill~\cite{liu2025speculative} & small model, 32-token chunks & 62.5 & 61.1 & 58.8 & 20.0 & 63.4 & 53.1 & \phantom{0}10\% & \phantom{0}1.0$\times$ \\
\rowcolor{ourrow} \cellcolor{white}\multirow{-6}{*}{\rotatebox[origin=c]{90}{\footnotesize Qwen3.5-9B}} & \method (ours) & Qwen3.5-0.8B & \textbf{64.0} & \textbf{75.0} & \textbf{87.0} & 18.3 & 69.4 & \textbf{62.7} & \phantom{0}10\% & \phantom{0}\textbf{3.7}$\times$ \\
\midrule
 & Full input & -- & 50.0 & 39.1 & \phantom{0}2.8 & 22.6 & 60.4 & 35.0 & 100\% & \phantom{0}1.0$\times$ \\
 & FastV~\cite{chen2024image} & target, one layer & \phantom{0}4.5 & 24.3 & \phantom{0}1.5 & 15.7 & \phantom{0}6.0 & 10.4 & \phantom{0}10\% & \phantom{0}3.5$\times$ \\
 & FlexSelect~\cite{zhang2025flexselect} & target, reference layer & 18.0 & 23.5 & \phantom{0}0.3 & 13.6 & \phantom{0}6.5 & 12.4 & \phantom{0}10\% & \phantom{0}2.4$\times$ \\
 & SGP~\cite{zhao2025stitch} & small model, same family & 47.0 & 32.4 & \phantom{0}3.5 & 19.0 & 37.5 & 27.9 & \phantom{0}10\% & \phantom{0}1.0$\times$ \\
 & SpecPrefill~\cite{liu2025speculative} & small model, 32-token chunks & 29.0 & 32.0 & \phantom{0}4.5 & \textbf{19.4} & 18.4 & 20.7 & \phantom{0}10\% & \phantom{0}1.0$\times$ \\
\rowcolor{ourrow} \cellcolor{white}\multirow{-6}{*}{\rotatebox[origin=c]{90}{\footnotesize Gemma-4-12B}} & \method (ours) & Qwen3.5-0.8B & \textbf{54.5} & \textbf{65.2} & \textbf{26.3} & 18.8 & \textbf{62.1} & \textbf{45.4} & \phantom{0}10\% & \textbf{11.6}$\times$ \\
\midrule
 & Full input & -- & 60.0 & 58.5 & 82.8 & 22.3 & 74.6 & 59.6 & 100\% & \phantom{0}1.0$\times$ \\
 & FastV~\cite{chen2024image} & target, one layer & 46.5 & 28.4 & 54.0 & 19.9 & 22.3 & 34.2 & \phantom{0}10\% & \phantom{0}4.0$\times$ \\
 & FlexSelect~\cite{zhang2025flexselect} & target, reference layer & 63.5 & 59.0 & \textbf{85.5} & 21.0 & 66.2 & 59.0 & \phantom{0}10\% & \phantom{0}1.1$\times$ \\
 & SGP~\cite{zhao2025stitch} & small model, same family & 57.0 & 58.9 & 84.5 & \textbf{21.7} & \textbf{74.5} & 59.3 & \phantom{0}10\% & \phantom{0}2.1$\times$ \\
 & SpecPrefill~\cite{liu2025speculative} & small model, 32-token chunks & 58.0 & 51.6 & 73.3 & 20.4 & 62.2 & 53.1 & \phantom{0}10\% & \phantom{0}2.3$\times$ \\
\rowcolor{ourrow} \cellcolor{white}\multirow{-6}{*}{\rotatebox[origin=c]{90}{\footnotesize Qwen3-VL-32B}} & \method (ours) & Qwen3.5-0.8B & \textbf{67.5} & \textbf{69.4} & 84.8 & 18.9 & 71.4 & \textbf{62.4} & \phantom{0}10\% & \phantom{0}\textbf{8.2}$\times$ \\
\bottomrule
\end{tabular}
\end{table*}

\subsection{Setup}
\label{sec:exp_setup}
We use Qwen3.5-9B~\cite{qwen2026qwen35}, Gemma-4-12B~\cite{google2026gemma4} and Qwen3-VL-32B-Thinking~\cite{bai2025qwen3vl}, each with a 256K-token context window, as target models and compare \method with four training-free pruning methods: FastV~\cite{chen2024image}, FlexSelect~\cite{zhang2025flexselect}, SGP~\cite{zhao2025stitch} and SpecPrefill~\cite{liu2025speculative}.
FastV and FlexSelect rank tokens by the attention of one layer of the target, SGP by that of a small model of the target's family that answers first (Qwen3.5-2B, Gemma-4-E2B and Qwen3-VL-2B), and SpecPrefill ranks 32-token chunks with the same small models.
For each question, FastV, FlexSelect and SGP keep as many single tokens as \method, images and text alike, and SpecPrefill keeps the top 10\% of its chunks; all four leave the kept tokens at their original positions.

We evaluate on MMLongBench~\cite{wang2025mmlongbench} at its longest length, 128K tokens, with 1,750 test inputs sampled from 19 datasets~\cite{chen2023infoseek,lerner2022viquae,wang2024needle,wu2025visual,krause2013cars,bossard2014food,vanhorn2021inat,xiao2010sun,huang2021govreport,shen2022multilexsum,deng2024longdocurl,ma2024mmlongbench,tanaka2023slidevqa} in five categories: visual RAG (VRAG), needle in a haystack (NIAH), many-shot in-context learning (ICL), summarization (Summ) and long-document VQA (DocQA).
We report MMLongBench's metric for each dataset (ROUGE-L F1 for summarization) and the mean of its category averages.

\method uses one setting for every target: Qwen3.5-0.8B as the selector, with images at half resolution, and head 7 of layer 7 and the threshold $\tau=-0.41$ from the synthetic documents of \cref{sec:skim}.
The primary modality receives 3/4 of the budget ($\alpha=3/4$), a value we did not tune, the last $W=256$ tokens of the document are always kept (\cref{sec:build}), and $Q$ holds the last 32 prompt tokens and the tokens of any question image.
Every target reasons with its recommended sampling settings for at most 32K tokens before it answers, and runs in SGLang~\cite{zheng2023sglang} on H100 GPUs, Qwen3-VL-32B on two with tensor parallelism.
Speed is the time to first token (TTFT) on 190 questions (10 per dataset) with batch size 1, image preprocessing and selection included; the selector runs on a separate H100, and on the target's GPUs it gives about the same speedups (see \cref{sec:suppl_speed}).
For more details on the benchmark, the sampling settings and the baselines, see \cref{sec:suppl_benchmark,sec:suppl_details}.

\subsection{Main results}
\label{sec:exp_main}

With 10\% of the input tokens (\cref{tab:main}), \method speeds up the prefill of every target by 3.7--11.6$\times$ and raises its average accuracy, the highest among the pruning methods on all three targets.
Both its speedup and its gain are largest on Gemma-4-12B (11.6$\times$; 35.0 to 45.4), whose full input is slow in SGLang and often reaches the 32K limit (\cref{sec:exp_speed}).
Where neither the full input nor \method reaches it, every target's mean score still rises by 3.3 to 13.4 points (see \cref{sec:suppl_accuracy}).
Without reasoning, \method stays 2.0 to 8.1 points above the full input on every target (see \cref{tab:nothink} in the appendix).

Accuracy rises most on needle questions, by 10.9 to 26.1 points, where the evidence is a few images or sentences, and in visual RAG, by 2.0 to 7.5 points.
But accuracy falls by 3.4 to 4.2 points in summarization, which needs the whole document. Long-document VQA also falls on Qwen3.5-9B and Qwen3-VL-32B, mainly on LongDocURL (\cref{sec:exp_failure}).

FlexSelect and SGP average between 0.9 points below and 1.1 above the full input on Qwen3.5-9B and Qwen3-VL-32B, at prefill speedups of 0.9--1.1$\times$ and 1.1--2.1$\times$, since FlexSelect runs most of the target's layers before it selects and SGP's small model writes an answer (\cref{sec:exp_speed}).
FastV is faster, by 1.6--4.0$\times$, but falls more than 20 points below the full input on every target.
SpecPrefill falls 6.5 to 14.3 points below the full input, at 1.0--2.3$\times$.

\begin{figure}[ht]
\centering
\includegraphics[width=\linewidth]{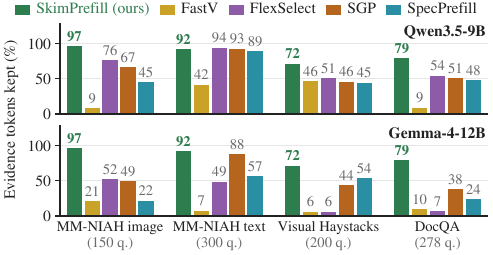}
\caption{Annotated evidence tokens kept at a 10\% budget (questions in parentheses).}
\label{fig:evidence}
\end{figure}
\subsection{Evidence kept by each method}
\label{sec:exp_evidence}
\Cref{fig:evidence} measures how much of the annotated evidence each method keeps with 10\% of the tokens.
\method keeps most of it, with one input for every target: 97\% of the needle-image tokens, 72\% of the Visual Haystacks needles and 79\% of the answer-page tokens.
FlexSelect and SGP keep 46--76\% of these image and page tokens on Qwen3.5-9B and less on Gemma-4-12B, where token-level selection leaves a few patches of most images and fragments of most paragraphs.
SpecPrefill keeps 45--48\% of them on Qwen3.5-9B and 22--54\% on Gemma-4-12B.
On Gemma-4-12B, FastV and FlexSelect keep 6--10\% of the Visual Haystacks needles and answer pages; FlexSelect's kept tokens spread evenly over the document.

\begin{figure}[t]
\centering
\includegraphics[width=\linewidth]{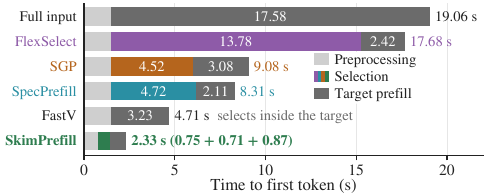}
\caption{Time to first token on Qwen3-VL-32B, split into preprocessing, selection and the target's prefill; 190 questions of about 105K input tokens (SGLang, batch size 1).}
\label{fig:speed}
\end{figure}

\subsection{Speed of each method}
\label{sec:exp_speed}
\begin{table}[t]
\centering
\caption{Generation length at 10\% of the input tokens (1,750 questions). \emph{Tokens}: mean generated tokens, reasoning and answer; \emph{32K}: answers that reach the 32K reasoning limit.}
\label{tab:generation}
\footnotesize
\setlength{\tabcolsep}{3.4pt}
\begin{tabular}{@{}lrrrrrr@{}}
\toprule
 & \multicolumn{2}{c}{Qwen3.5-9B} & \multicolumn{2}{c}{Gemma-4-12B} & \multicolumn{2}{c}{Qwen3-VL-32B} \\
\cmidrule(l{2pt}r{2pt}){2-3}\cmidrule(l{2pt}r{2pt}){4-5}\cmidrule(l{2pt}r{2pt}){6-7}
Method & Tokens & 32K & Tokens & 32K & Tokens & 32K \\
\midrule
Full input & 6.4K & 102 & 10.4K & 435 & 1.9K & 50 \\
FastV & 9.1K & 204 & 18.9K & 875 & 5.4K & 218 \\
FlexSelect & 5.9K & 105 & 22.5K & 1,007 & 4.1K & 169 \\
SGP & 6.3K & 115 & 14.3K & 626 & 3.9K & 158 \\
SpecPrefill & 6.5K & 106 & 24.0K & 1,175 & 2.9K & 98 \\
\rowcolor{ourrow} \method & 3.5K & 15 & 7.3K & 266 & 1.5K & 12 \\
\bottomrule
\end{tabular}
\end{table}

On Qwen3-VL-32B (\cref{fig:speed}), \method waits 2.33\,s for the first token, against 19.06\,s for the full input, since its selector stops at layer 7 of a 0.8B model and the target prefills only the kept units.
The baselines pay more for their selection: FlexSelect runs 49 of the target's 64 layers over all 105K tokens (13.78\,s), the small models of SGP and SpecPrefill read the whole input and both generate some tokens (4.52 and 4.72\,s), and FastV, which selects inside the target, waits 4.71\,s in all.
On all three targets, preprocessing and selection take 1.35--1.46\,s, so the speedups of \cref{tab:main} grow with the target's prefill time, up to 11.6$\times$ on Gemma-4-12B, whose full input takes 34.05\,s.
For more details, see \cref{tab:answer} in the appendix.

\paragraph{Generation length.}
Every target answers in reasoning mode, so the wait for the answer also depends on how long it thinks.
In \cref{tab:generation}, every target generates fewer tokens with \method than with the full input and reaches the 32K limit less often, whereas token-level and block-level selection (SpecPrefill) make most targets think longer; \cref{sec:suppl_accuracy} analyzes the answers of Gemma-4-12B that reach this limit.
From the request to the last answer token, \method answers 2.2--3.8$\times$ sooner than the full input, against 0.6--1.6$\times$ for FastV, FlexSelect, SGP and SpecPrefill.
\Cref{tab:answer} in the appendix splits this time into prefill and decoding.

\subsection{Budget and input length}
\label{sec:exp_budget}
\label{sec:exp_length}
\begin{figure}[t]
\centering
\includegraphics[width=\linewidth]{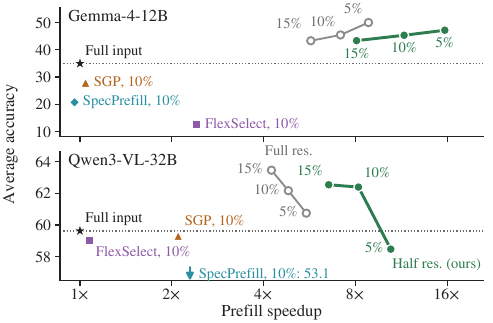}
\caption{Accuracy against prefill speedup at budgets of 5, 10 and 15\% (labels), with the selector at half (ours) or full resolution.}
\label{fig:budget}
\end{figure}
\begin{figure}[t]
\centering
\includegraphics[width=\linewidth]{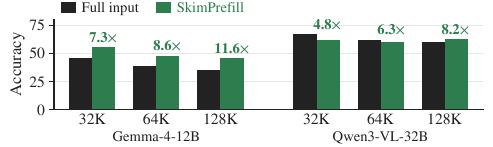}
\caption{Accuracy at three input lengths with 10\% of the input tokens kept; above the bars of \method: its prefill speedup.}
\label{fig:length}
\end{figure}
A larger budget helps Qwen3-VL-32B and hurts Gemma-4-12B (\cref{fig:budget}).
From 5\% to 15\%, the average of Qwen3-VL-32B rises by 4.0 points, mostly in long-document VQA (59.6 to 71.9; \cref{tab:budget_categories}), where more pages hold more of the evidence, and that of Gemma-4-12B falls by 3.8 points but stays above the full input at every budget.
Its fall comes mostly from many-shot ICL, where its score drops from 46.0 to 18.3 (see \cref{sec:suppl_failure}).

On the same questions at 32K and 64K tokens (\cref{fig:length}), the speedup grows with the input length on both targets, as the target's prefill grows faster than the selector's.
\method stays above the full input at every length on Gemma-4-12B, and on Qwen3-VL-32B it falls 5.1 and 1.5 points below it at 32K and 64K, mostly in long-document VQA (60.5 against 81.3 at 32K, where 10\% is about 3K tokens).

\subsection{Ablations}
\label{sec:exp_ablation}
\paragraph{Selector resolution.}
\label{sec:exp_resolution}
The selector reads every image at half resolution, since it only has to find the units that hold the evidence.
At a 10\% budget, reading the original images changes the average by less than one point on every target (see \cref{tab:resolution} in the appendix), while halving them raises the speedup by 1.6--1.8$\times$.

\paragraph{Whole units and a good ranking are both needed.}
With the same selector scores (\cref{tab:ablation}), keeping the highest-scoring tokens instead of whole units falls far below the full input on both targets (37.9 and 15.4) and makes both targets think longer than with whole units, and evenly spaced whole units stay below it (38.7 and 25.7).
Whole units matter more than their positions: tokens stay below the full input at consecutive positions, and whole units stay above it at their original positions.
The mean of all layer-7 heads ranges from 0.8 points below to 0.1 above \method, while a random head falls to 50.2 and 34.6.

We also apply our whole-unit selection to the scores of SGP and FlexSelect; it raises their accuracy over single tokens (\cref{tab:main,tab:ablation}), and on Qwen3.5-9B \method is 1.3 and 1.2 points below them at a much lower cost.
Scoring with every full-attention head of the selector, following SpecPrefill, is less accurate and slower than \method on both targets (53.0 against 62.7 on Qwen3.5-9B).
Larger selectors raise the average by up to 0.6 points but lower the speedup: the full-attention Qwen3-VL-2B lowers it more than the hybrid Qwen3.5-2B, although both run 12 layers (see \cref{sec:suppl_ablation}).
\begin{table}[t]
\centering
\caption{Ablations at 128K tokens, 10\% kept (speedup as in \cref{tab:main}; --: not measured); \method keeps whole units at consecutive positions, and the last three rows rank all units in one list.}
\label{tab:ablation}
\footnotesize
\setlength{\tabcolsep}{3.4pt}
\begin{tabular}{lrrrr}
\toprule
 & \multicolumn{2}{c}{Qwen3.5-9B} & \multicolumn{2}{c}{Gemma-4-12B} \\
\cmidrule(lr){2-3}\cmidrule(lr){4-5}
Variant & Acc. & Speedup & Acc. & Speedup \\
\midrule
Full input & 60.5 & 1.0$\times$ & 35.0 & 1.0$\times$ \\
\rowcolor{ourrow} \method & 62.7 & 3.7$\times$ & 45.4 & 11.6$\times$ \\
\midrule
\multicolumn{5}{l}{\textit{What is kept}} \\
\quad Top tokens, original positions & 37.9 & \multicolumn{1}{c}{--} & 15.4 & \multicolumn{1}{c}{--} \\
\quad Top tokens, consecutive positions & 38.8 & \multicolumn{1}{c}{--} & 25.5 & \multicolumn{1}{c}{--} \\
\quad Whole units, original positions & 61.6 & \multicolumn{1}{c}{--} & 37.8 & \multicolumn{1}{c}{--} \\
\quad Evenly spaced units & 38.7 & \multicolumn{1}{c}{--} & 25.7 & \multicolumn{1}{c}{--} \\
\addlinespace[3pt]
\multicolumn{5}{l}{\textit{Whole units of \method, ranked by}} \\
\quad Mean of all heads of layer 7 & 61.9 & \multicolumn{1}{c}{--} & 45.5 & \multicolumn{1}{c}{--} \\
\quad Random head of layer 7 & 50.2 & \multicolumn{1}{c}{--} & 34.6 & \multicolumn{1}{c}{--} \\
\quad SGP's scores & 64.0 & 1.5$\times$ & 44.3 & 1.2$\times$ \\
\quad FlexSelect's scores & 63.9 & 1.2$\times$ & 30.9 & 3.3$\times$ \\
\quad All layers, as in SpecPrefill & 53.0 & 3.1$\times$ & 38.4 & 10.0$\times$ \\
\bottomrule
\end{tabular}
\end{table}

\begin{figure}[h]
\centering
\includegraphics[width=\linewidth]{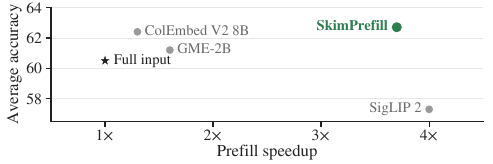}
\caption{Average accuracy against prefill speedup when a retriever replaces the head (Qwen3.5-9B, 10\% kept); for SigLIP 2 and ColEmbed V2 8B, BM25~\cite{robertson2009bm25} ranks the paragraphs.}
\label{fig:retrieval}
\end{figure}

\paragraph{Retrievers as scorers.}
A retriever can rank the units of \method in place of the head, each unit against the question on its own (\cref{fig:retrieval}).
SigLIP 2~\cite{tschannen2025siglip2} is about as fast but 5.4 points less accurate; ColEmbed V2 8B~\cite{moreira2026colembed} and GME-2B~\cite{zhang2025gme}, which encode each image separately with a larger model, fall 0.3 and 1.5 points below \method, at only 1.3$\times$ and 1.6$\times$.

\subsection{Failure analysis}
\label{sec:exp_failure}
\begin{table}[h]
\centering
\caption{\method on the 100 LongDocURL questions, whose documents at 128K tokens have 57 pages on average. \emph{Pages}: pages kept; \emph{Missed}: questions with no answer page kept.}
\label{tab:longdocurl}
\footnotesize
\setlength{\tabcolsep}{3.4pt}
\begin{tabular}{lrrcc}
\toprule
 & & & \multicolumn{2}{c}{Accuracy} \\
\cmidrule(l){4-5}
Method & Pages & Missed & Qwen3-VL-32B & Gemma-4-12B \\
\midrule
Full input & 57.0 & 0 & 76.2 & 57.8 \\
SGP, 10\% &  &  & 73.9 & 24.3 \\
\midrule
\method, 5\% & 2.6 & 39 & 45.8 & 40.2 \\
\rowcolor{ourrow} \method, 10\% & 6.1 & 18 & 62.5 & 53.0 \\
\method, 15\% & 9.5 & 12 & 66.0 & 56.1 \\
\bottomrule
\end{tabular}
\end{table}

\paragraph{\method on long documents.}
\method falls behind SGP on LongDocURL, by 11.4 points on Qwen3-VL-32B and 14.5 on Qwen3.5-9B.
For 18 of the 100 questions, none of the pages that \method keeps holds the answer (\cref{tab:longdocurl}), while SGP keeps a few tokens on nearly every page and answers most of these questions.
If \method scored as well as SGP on these 18, its gap would shrink from 11.4 to 0.8 points, so the gap lies in the page ranking (see \cref{sec:suppl_failure}).
A 15\% budget keeps an answer page for 6 of the 18 questions, but \method stays 7.9 points below SGP at 10\% on Qwen3-VL-32B.

\begin{figure}[t]
\centering
\includegraphics[width=\linewidth]{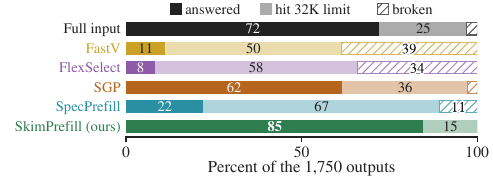}
\caption{How the 1,750 outputs of Gemma-4-12B end with 10\% of the input tokens kept; a broken output opens the reasoning tag again or ends without an answer.}
\label{fig:failure}
\end{figure}
\paragraph{Baselines on Gemma-4-12B.}
On Gemma-4-12B, all four baselines fall below the full input (\cref{tab:main}).
In \cref{fig:failure}, only 11\% and 8\% of the outputs with FastV and FlexSelect give an answer (22\% with SpecPrefill), against 72\% with the full input and 85\% with \method.
Keeping single tokens explains most of the drop: with the same scores, whole units raise FlexSelect by 18.5 points and SGP by 16.4 (\cref{tab:main,tab:ablation}).
\Cref{tab:ablation} also shows that the head of \method in Qwen3.5-0.8B, from another family, gives about the same accuracy as SGP's Gemma-4-E2B, a small model of the target's family (45.4 and 44.3), at 11.6$\times$ instead of 1.2$\times$.
In many-shot ICL, token-level selection also splits the examples apart (see \cref{sec:suppl_failure}).

\section{Conclusion}
\label{sec:conclusion}
We presented \method, which builds on the finding that the evidence for a question lies in a few whole units.
A small hybrid model reads the input up to an early layer and ranks its units with one attention head of that layer; the target then prefills only the top-ranked units.
With a Qwen3.5-0.8B selector that needs no training, \method keeps 10\% of the input tokens and speeds up the prefill of three target models on MMLongBench at 128K tokens by 3.7--11.6$\times$ while maintaining their average accuracy.
\section*{Acknowledgement}

This work is funded in part by Federal Ministry of Research, Technology and Space of Germany (BMFTR) - project ID 16IS26046 and by the European Union - European Research Council (ERC) Starting Grant -
Project-ID 101219243. Views and opinions expressed are however those of the author(s) only and do
not necessarily reflect those of the European Union or the European Research Council. Neither the
European Union nor the granting authority can be held responsible for them.

{
    \small
    \bibliographystyle{ieeenat_fullname}
    \bibliography{references}
}

\clearpage
\appendix
\crefalias{section}{appendix}
\crefalias{subsection}{subappendix}

\section{Benchmark}
\label{sec:suppl_benchmark}

\paragraph{Datasets.}
We sample 1,750 test inputs from MMLongBench at 128K tokens.
They come from 19 datasets in five categories: visual RAG (InfoSeek~\cite{chen2023infoseek} and ViQuAE~\cite{lerner2022viquae}), needle in a haystack (the text and image versions of three MM-NIAH tasks~\cite{wang2024needle} and two Visual Haystacks tasks~\cite{wu2025visual}), many-shot in-context learning (Cars196~\cite{krause2013cars}, Food101~\cite{bossard2014food}, iNat2021~\cite{vanhorn2021inat} and SUN397~\cite{xiao2010sun}), summarization (GovReport~\cite{huang2021govreport} and Multi-LexSum~\cite{shen2022multilexsum}) and long-document VQA (LongDocURL~\cite{deng2024longdocurl}, MMLongBench-Doc~\cite{ma2024mmlongbench} and SlideVQA~\cite{tanaka2023slidevqa}).
MMLongBench's overview table lists 16 datasets, since it counts the two versions of each MM-NIAH task as one.
Each MM-NIAH image task has 50 inputs, built from 18 to 26 distinct questions, each MM-NIAH text task 100 questions (MMLongBench itself samples 50), and every other dataset 100 questions.
Datasets that place the evidence at several depths of the input contribute one depth per question; on Visual Haystacks (single), MM-NIAH retrieval (text) and ViQuAE, this depth is the start of the input.

\paragraph{Scoring.}
As in MMLongBench, the two versions of each MM-NIAH task are averaged first, so the needle category averages five scores: the three MM-NIAH tasks and the two Visual Haystacks tasks.
MM-NIAH counting uses MMLongBench's metric, which compares the sum of the listed counts with the sum of the annotated counts instead of MM-NIAH's position-wise soft accuracy, so dropping an image without a needle does not change the score.
For summarization, we report ROUGE-L F1 in place of MMLongBench's GPT-4o-based score.

\section{Implementation details}
\label{sec:suppl_details}

\subsection{Targets and generation}
\label{sec:suppl_targets}

\paragraph{Targets.}
Qwen3.5-9B~\cite{qwen2026qwen35} combines Gated DeltaNet layers with full-attention layers, Gemma-4-12B~\cite{google2026gemma4} combines sliding-window layers with global attention layers, and Qwen3-VL-32B-Thinking~\cite{bai2025qwen3vl} uses full attention in all layers, according to its released configuration.

\paragraph{Reasoning mode.}
Every target answers in its reasoning mode, in SGLang for every method, and all methods use the same sampling seed for a given question.
For accuracy, the targets run with SGLang's deterministic inference.
Qwen3.5-9B and Gemma-4-12B turn on the thinking mode of their own chat templates, and Qwen3-VL-32B-Thinking always reasons.
Sampling uses the recommended values: temperature 1.0 and top-p 0.95, and top-k 64 for Gemma-4-12B and 20 for the Qwen targets.
The reasoning may take up to 32,768 tokens.
If it has not ended by then, we append the end-of-reasoning tag: \texttt{</think>} for the Qwen targets and \texttt{<channel|>} for Gemma-4-12B, which reasons between \texttt{<|channel>thought} and \texttt{<channel|>}.
The Qwen targets then write their answer in at most 512 tokens (1,536 for summarization), and Gemma-4-12B, after a forced tag, in at most 1,024.
Only the answer is scored.
We also run every target without reasoning (\cref{sec:suppl_accuracy}).

\subsection{Experimental settings}
\label{sec:suppl_settings}
\paragraph{Implementation.}
Every target answers in SGLang with the prompts, sampling settings, reasoning limits and seeds of \cref{sec:suppl_targets}.
For accuracy, the scores that choose the kept tokens are computed in PyTorch for the selector and for the baselines, except FastV, which ranks the tokens inside the target in SGLang.
For timing, every step runs in SGLang, except the retrievers of \cref{sec:suppl_ablation}, which are timed in their scoring runs.
The selector of \method halves the image size with PIL's bicubic filter on 16 CPU threads, and this step counts as preprocessing.

\paragraph{Units.}
Text is split into paragraph units at blank lines and at images.
An image unit is an image together with the short label lines next to it (\cref{sec:prelim}).
A piece of text with fewer than 8 tokens is not a unit; it is kept only as part of an image's label or of the last $W$ tokens.
The score $s_i$ of an image unit (\cref{sec:skim}) uses only its visual tokens, not its label.

\paragraph{Baseline settings.}
In their papers, FastV, FlexSelect and SGP rank only visual tokens and SpecPrefill only text; here all four rank every document token, images and text alike.
FastV, FlexSelect and SGP keep the instruction and the question and fill the rest with document tokens, so that the target prefills as many tokens as with \method (about 10\% of the document).
SpecPrefill keeps the top 10\% of its chunks, and the four baselines leave the kept tokens at their original positions.

\subsection{Baselines}
\label{sec:suppl_baselines}
We compare only with scorers that need no training, as \method needs none, and leave out IF-Prune~\cite{sun2026ifprune}, which trains a small model that must share the target's architecture, CAPS~\cite{zhu2026attention}, which fine-tunes a VLM to rank pages, and FlexSelect's lightweight selector~\cite{zhang2025flexselect}, which is trained to mimic the ranking of the target's reference layer.
\paragraph{FlexSelect.}
We follow the training-free form of FlexSelect~\cite{zhang2025flexselect} and its official code: at one reference layer of the target, each document token scores the largest attention it receives from the tokens after the document, over all heads, and the whole document is ranked at once, with no frame groups.
We choose this layer with a needle test, as FlexSelect does, on 100 MM-NIAH image questions outside the test set: among the layers that attend to the whole input, we take the one whose top 10\% of tokens holds the most needle-image tokens, namely layer 19 of Qwen3.5-9B (recall 0.66), layer 5 of Gemma-4-12B (0.50; layer 29: 0.28) and layer 49 of Qwen3-VL-32B (0.69).
In \cref{fig:finding}, where one head ranks whole images, the best layers are 29 and 11.

\paragraph{SGP.}
SGP is the token-pruning part of SGL~\cite{zhao2025stitch} without its early exit, and we follow its official code: the small model of the target's family answers first, greedily, each document token scores the attention it receives from the question and the generated tokens, summed over heads, steps and layers, and the target drops the low-scoring tokens at layer 2, as in SGL's lowest-retention setting.
We let the small model answer without reasoning in at most 64 tokens on every dataset (SGL's code sets 10 or 100 per dataset).

\paragraph{FastV.}
FastV~\cite{chen2024image} ranks the tokens at one layer by the attention of the last prompt token, averaged over heads, as in its official code (the paper uses the average attention a token receives), and the target drops the rest from the next layer on.
We rank once on the prompt, as in its KV-cache version.
Qwen3-VL-32B ranks at layer 1 (FastV's $K=2$); Qwen3.5-9B and Gemma-4-12B rank at their first layer that attends to the whole input, layer 3 after Gated DeltaNet layers and layer 5 after sliding-window layers.

\paragraph{SpecPrefill.}
The released code of SpecPrefill~\cite{liu2025speculative} has speculators only for Llama targets, so we use SGP's small models (Qwen3.5-2B, Gemma-4-E2B and Qwen3-VL-2B) and take the attention of their full-attention layers only: the six of Qwen3.5-2B, the global layers of Gemma-4-E2B and all layers of Qwen3-VL-2B.
As in the released code, the last prompt token and up to 7 generated tokens attend to the prompt, the attention is smoothed over 13 tokens, maximized over layers and heads and averaged over these queries, and the target reads the top 10\% of the chunks of 32 consecutive tokens.
We generate these tokens greedily, as the SpecPrefill paper does in its evaluations, since sampling them at the target's temperature of 1 would make the selection random.
We chunk the image tokens, at full resolution, in prompt order like text, because the released code handles only text.

\paragraph{Retrievers.}
In \cref{fig:retrieval}, three retrievers, SigLIP~2, ColEmbed~V2 8B and GME-2B, replace the head of \method; only the unit scores change.
They rank the units of \method, a full-resolution image with its label lines or a paragraph, and fill the same token budget.
Their image and text scores are not on one scale, so each question keeps the primary modality that \method chose for it, and with it the split of the budget.
Their times leave out the selector's pass that makes this choice, which favors them.
Each query is the question part of the prompt with its images.

SigLIP~2 (so400m, 384 pixels) scores an image unit by the higher of its cosine similarities with the question text and with each question image, and reciprocal rank fusion~\cite{cormack2009rrf} ($k=60$) combines this ranking with the BM25 ranking of the unit's label against the question text.
ColEmbed~V2 encodes an image unit, the image with its label, as one multi-vector document and scores it against each query separately: every query token takes its highest similarity over the document's tokens, and these are averaged over the query's tokens.
The queries are the question text and each question image, encoded as an image document; the unit keeps the highest of these scores.

Since ColEmbed~V2 accepts a document only with an image and SigLIP~2 reads only 64 text tokens, BM25~\cite{robertson2009bm25} ranks the paragraphs against the question text for both.
In VRAG, where the question image is the only image, both also rank the passages against this image and fuse this ranking with that of BM25 in the same way: SigLIP~2 by the cosine similarity of a passage's first 64 tokens with the image, ColEmbed~V2 by late interaction of the passage, encoded as a query of up to 512 tokens, with the image.

GME-2B embeds every unit as one vector, and the question once with each of its images under the instruction ``Find an image or a passage that helps answer the question.''; each unit is ranked by its highest cosine similarity with these query vectors.

\subsection{\texorpdfstring{Variants in \cref{tab:ablation}}{Variants in Table 4}}
\label{sec:suppl_variants}
\paragraph{Positions.}
Top tokens at original positions keep each token at its position in the full prompt; at consecutive positions, the kept tokens are numbered one after another.
Whole units at original positions give the target \method's short prompt with each token at its position in the full prompt.
The blank line that \method inserts between two kept units takes the position right after the last token of the unit before it.

\paragraph{SpecPrefill's scoring.}
The row ``All layers, as in SpecPrefill'' of \cref{tab:ablation} replaces the head of \method with SpecPrefill's scoring (\cref{sec:suppl_baselines}): the selector reads the images at half resolution, runs all 24 layers and scores each token by the attention that the last prompt token and up to 7 greedy generated tokens pay to it in its 6 full-attention layers.
A unit's score is the mean over its tokens, and all units are ranked in one list, as in the rows with SGP's and FlexSelect's scores.
Running all 24 layers lowers the speedup from 3.7$\times$ to 3.1$\times$ on Qwen3.5-9B and from 11.6$\times$ to 10.0$\times$ on Gemma-4-12B (\cref{tab:ablation}).

\section{Calibration documents}
\label{sec:suppl_selector}
\label{sec:suppl_calib}

\paragraph{Calibration data construction.}
We choose the selector's head and its threshold $\tau$ (\cref{sec:skim,sec:build}) on 50 synthetic documents that share no text or image with MMLongBench.
Each document starts with the instruction ``Read the document below, which mixes text and pictures, and answer the question at the end.'' and ends with a question.
Between them are consecutive paragraphs of PG-19~\cite{rae2020compressive} books, 393 to 2,701 per document, and 64 pictures from nine MileBench~\cite{song2024milebench} subsets (nuScenes, ALFRED, TQA and six video-frame subsets) at random paragraph boundaries.
Paragraphs are added until the document reaches about 108K Qwen3.5 tokens (107,601 to 108,064), close to the MM-NIAH image documents of the 128K setting (108K to 111K tokens).
Every paragraph and picture is used once in the whole set, a document holds at most one frame of a video, and pictures are shrunk to a long side of at most 768 pixels.
A paragraph is dropped when any of its 13-word windows occurs in an MMLongBench data file (43 of 109,723 paragraphs).
A picture is dropped when it equals or nearly duplicates one of the 269,945 MMLongBench images; none did.

\paragraph{Needle construction.}
Each document carries one image needle, a book cover from the OCR-VQA subset of MileBench, in one of three kinds: pasted into one picture (17 documents), pasted into two or three pictures (17) or inserted as a picture of its own (16).
For the first and the last kind, the question shows four covers and asks which one is in the document; for the second, it asks how many copies each picture holds.
Each cover, as needle or option, is used once (149 in all).
A second copy of each document drops the image question and its option images and adds text needles, sentences appended to randomly chosen paragraphs, with a question about them.
The text needle is a fact asked in other words (17 documents), two to four counts to be listed (17) or two facts to be combined (16).
An example of the first kind is ``The choir of Tallinn rehearses in a converted grain store.'', asked as ``In what kind of building does the singing group of Tallinn practise?''
The needle sentences are our own and share no 6-word window with MMLongBench.

\begin{table}[h]
\centering
\caption{Heads of the selector on the calibration questions: for each full-attention layer, the head whose three highest-scoring units hold a needle for the most questions. \emph{Image}, \emph{Text}: how many of the 50 image and the 50 text questions they are; \emph{Layers run}: layers the selector runs when it stops after this one.}
\label{tab:calib_heads}
\footnotesize
\begin{tabular}{@{}cccrr@{}}
\toprule
Layer & Head & Layers run & Image & Text \\
\midrule
3 & 7 & 4 & 49 & 25 \\
\rowcolor{ourrow} 7 & 7 & 8 & \textbf{50} & 31 \\
11 & 3 & 12 & 47 & \textbf{38} \\
15 & 2 & 16 & 46 & 34 \\
19 & 5 & 20 & 47 & 31 \\
23 & 3 & 24 & 47 & 27 \\
\bottomrule
\end{tabular}
\end{table}

\paragraph{Head and threshold selection.}
The selector scores each unit by the attention that the question pays to it in one head of one layer (\cref{sec:skim}).
To choose this head, we leave out the text units within 256 tokens of the question and count, for each head, the questions whose three highest-scoring units hold the needle.
The head with the most image questions wins; ties go to more text questions and then to the earlier layer.
\Cref{tab:calib_heads} gives, for each full-attention layer of the selector, the head whose three highest-scoring units hold a needle for the most questions.
We use head 7 of layer 7.
Its top three units hold the needle for all 50 image questions, which no other head reaches, and for 31 of the 50 text questions, against 25 to 38 for the other layers.
This head is not the best one for text needles: layer 11 finds more of them, but the selector would then run 12 layers instead of 8, so we trade these few text needles for a shorter selector pass.
For $\tau$, each calibration question gives its $\Delta$ of \cref{eq:primary}.
On the calibration questions that set it, $\tau=-0.4115$ best separates the image questions ($\Delta\ge\tau$) from the text questions, with a balanced accuracy of 0.99 (median $\Delta$: 0.35 for the image and $-1.55$ for the text questions).

\paragraph{Code and data release.}
We will release our code, with the scripts and configurations that reproduce our experiments.
It will also include the scripts that rebuild the calibration documents from the public PG-19 and MileBench releases, downloaded at fixed revisions, and the checksums of all 3,427 files of our set, against which a rebuild is checked.
We do not redistribute the pictures and the covers, whose source datasets have their own license terms.

\begin{table*}[t]
\centering
\caption{Accuracy on each of the 19 datasets of MMLongBench~\cite{wang2025mmlongbench} at 128K input tokens, with 10\% of the input tokens kept; 100 test inputs per dataset, 50 for the three MM-NIAH image tasks. Each dataset uses MMLongBench's metric (summarization: ROUGE-L F1); the average is that of \cref{tab:main}. NIAH: MM-NIAH; \emph{Flex.}: FlexSelect; \emph{Spec.}: SpecPrefill. \textbf{Bold}: best of the pruning methods for each target.}
\label{tab:datasets}
\footnotesize
\setlength{\tabcolsep}{1.5pt}
\ifdefined\pdcolw\else\newlength{\pdcolw}\fi\settowidth{\pdcolw}{FastV}
\begin{tabular}{l>{\raggedleft\arraybackslash}p{\pdcolw}>{\raggedleft\arraybackslash}p{\pdcolw}>{\raggedleft\arraybackslash}p{\pdcolw}>{\raggedleft\arraybackslash}p{\pdcolw}>{\raggedleft\arraybackslash}p{\pdcolw}>{\columncolor{ourrow}\raggedleft\arraybackslash}p{\pdcolw}@{\hspace{9pt}}c@{}>{\raggedleft\arraybackslash}p{\pdcolw}>{\raggedleft\arraybackslash}p{\pdcolw}>{\raggedleft\arraybackslash}p{\pdcolw}>{\raggedleft\arraybackslash}p{\pdcolw}>{\raggedleft\arraybackslash}p{\pdcolw}>{\columncolor{ourrow}\raggedleft\arraybackslash}p{\pdcolw}@{\hspace{9pt}}c@{}>{\raggedleft\arraybackslash}p{\pdcolw}>{\raggedleft\arraybackslash}p{\pdcolw}>{\raggedleft\arraybackslash}p{\pdcolw}>{\raggedleft\arraybackslash}p{\pdcolw}>{\raggedleft\arraybackslash}p{\pdcolw}>{\columncolor{ourrow}\raggedleft\arraybackslash}p{\pdcolw}}
\toprule
 & \multicolumn{6}{c}{Qwen3.5-9B} & & \multicolumn{6}{c}{Gemma-4-12B} & & \multicolumn{6}{c}{Qwen3-VL-32B} \\
\cmidrule(lr){2-7}\cmidrule(lr){9-14}\cmidrule(lr){16-21}
Dataset & \multicolumn{1}{c}{Full} & \multicolumn{1}{c}{FastV} & \multicolumn{1}{c}{Flex.} & \multicolumn{1}{c}{SGP} & \multicolumn{1}{c}{Spec.} & \multicolumn{1}{>{\columncolor{ourrow}}c}{Ours} & & \multicolumn{1}{c}{Full} & \multicolumn{1}{c}{FastV} & \multicolumn{1}{c}{Flex.} & \multicolumn{1}{c}{SGP} & \multicolumn{1}{c}{Spec.} & \multicolumn{1}{>{\columncolor{ourrow}}c}{Ours} & & \multicolumn{1}{c}{Full} & \multicolumn{1}{c}{FastV} & \multicolumn{1}{c}{Flex.} & \multicolumn{1}{c}{SGP} & \multicolumn{1}{c}{Spec.} & \multicolumn{1}{>{\columncolor{ourrow}}c}{Ours} \\
\midrule
\multicolumn{21}{@{}l}{\textit{Visual RAG}} \\
\quad InfoSeek & 53.0 & 40.0 & 50.0 & \textbf{52.0} & 51.0 & \textbf{52.0} & & 40.0 & 5.0 & 12.0 & 25.0 & 19.0 & \textbf{43.0} & & 50.0 & 35.0 & 53.0 & 45.0 & 42.0 & \textbf{54.0} \\
\quad ViQuAE & 71.0 & \textbf{82.0} & 74.0 & 69.0 & 74.0 & 76.0 & & 60.0 & 4.0 & 24.0 & \textbf{69.0} & 39.0 & 66.0 & & 70.0 & 58.0 & 74.0 & 69.0 & 74.0 & \textbf{81.0} \\
\multicolumn{21}{@{}l}{\textit{Needle in a haystack}} \\
\quad NIAH retrieval, image & 60.0 & 44.0 & 80.0 & 74.0 & 70.0 & \textbf{88.0} & & 14.0 & 14.0 & 28.0 & 18.0 & 24.0 & \textbf{68.0} & & 72.0 & 26.0 & 78.0 & \textbf{90.0} & 42.0 & 78.0 \\
\quad NIAH counting, image & 20.0 & 20.0 & 46.0 & 32.0 & 34.0 & \textbf{56.0} & & 6.0 & 0.0 & 2.0 & 8.0 & 8.0 & \textbf{34.0} & & 16.0 & 26.0 & 14.0 & 16.0 & 22.0 & \textbf{32.0} \\
\quad NIAH reasoning, image & 48.0 & 42.0 & 60.0 & 58.0 & 44.0 & \textbf{70.0} & & 32.0 & 14.0 & 26.0 & 36.0 & 32.0 & \textbf{48.0} & & 50.0 & 30.0 & \textbf{48.0} & 44.0 & 40.0 & \textbf{48.0} \\
\quad NIAH retrieval, text & 100.0 & 83.0 & 99.0 & \textbf{100.0} & 97.0 & 99.0 & & 87.0 & 1.0 & 3.0 & 48.0 & 15.0 & \textbf{100.0} & & 100.0 & 6.0 & 95.0 & 96.0 & \textbf{99.0} & 98.0 \\
\quad NIAH counting, text & 60.0 & 1.0 & 41.0 & 26.0 & 46.0 & \textbf{86.0} & & 9.0 & 0.0 & 0.0 & 4.0 & 3.0 & \textbf{87.0} & & 41.0 & 3.0 & 48.0 & 49.0 & 42.0 & \textbf{88.0} \\
\quad NIAH reasoning, text & 69.0 & 6.0 & 53.0 & 43.0 & 62.0 & \textbf{71.0} & & 33.0 & 0.0 & 0.0 & 4.0 & 2.0 & \textbf{77.0} & & 56.0 & 5.0 & 43.0 & 48.0 & 43.0 & \textbf{74.0} \\
\quad Visual Haystacks, single & 58.0 & 67.0 & 64.0 & 65.0 & 69.0 & \textbf{72.0} & & 57.0 & 55.0 & 47.0 & 56.0 & 63.0 & \textbf{64.0} & & 64.0 & 46.0 & 63.0 & 64.0 & \textbf{67.0} & \textbf{67.0} \\
\quad Visual Haystacks, multi & 66.0 & 45.0 & \textbf{70.0} & 67.0 & 60.0 & 68.0 & & 48.0 & 52.0 & 41.0 & 47.0 & \textbf{55.0} & \textbf{55.0} & & 61.0 & 48.0 & 69.0 & 59.0 & 47.0 & \textbf{71.0} \\
\multicolumn{21}{@{}l}{\textit{Many-shot ICL}} \\
\quad Cars196 & 95.0 & 24.0 & 95.0 & 89.0 & 57.0 & \textbf{96.0} & & 2.0 & 1.0 & 0.0 & 3.0 & 6.0 & \textbf{22.0} & & 85.0 & 48.0 & 89.0 & \textbf{90.0} & 85.0 & \textbf{90.0} \\
\quad Food101 & 84.0 & 65.0 & \textbf{86.0} & 85.0 & 59.0 & \textbf{86.0} & & 3.0 & 0.0 & 0.0 & 4.0 & 5.0 & \textbf{33.0} & & 80.0 & 55.0 & 79.0 & 79.0 & 61.0 & \textbf{81.0} \\
\quad iNat2021 & 79.0 & 51.0 & 78.0 & 78.0 & 46.0 & \textbf{81.0} & & 2.0 & 2.0 & 1.0 & 4.0 & 4.0 & \textbf{16.0} & & 80.0 & 42.0 & \textbf{87.0} & 86.0 & 61.0 & 80.0 \\
\quad SUN397 & 84.0 & 63.0 & 83.0 & 83.0 & 73.0 & \textbf{85.0} & & 4.0 & 3.0 & 0.0 & 3.0 & 3.0 & \textbf{34.0} & & 86.0 & 71.0 & 87.0 & 83.0 & 86.0 & \textbf{88.0} \\
\multicolumn{21}{@{}l}{\textit{Summarization}} \\
\quad GovReport & 23.0 & 16.9 & 20.0 & \textbf{20.9} & 18.5 & 17.8 & & 23.3 & 14.9 & 12.5 & 17.4 & 16.6 & \textbf{18.6} & & 22.6 & 19.6 & 19.8 & \textbf{21.4} & 19.4 & 18.4 \\
\quad Multi-LexSum & 21.9 & 20.0 & 21.3 & 21.0 & \textbf{21.4} & 18.7 & & 21.8 & 16.5 & 14.6 & 20.6 & \textbf{22.3} & 19.0 & & 22.0 & 20.1 & \textbf{22.2} & 21.9 & 21.4 & 19.4 \\
\multicolumn{21}{@{}l}{\textit{Long-document VQA}} \\
\quad LongDocURL & 74.9 & 16.0 & 66.3 & \textbf{73.8} & 56.4 & 59.3 & & 57.8 & 5.9 & 5.9 & 24.3 & 11.2 & \textbf{53.0} & & 76.2 & 17.9 & 59.0 & \textbf{73.9} & 57.0 & 62.5 \\
\quad MMLongBench-Doc & 57.1 & 27.7 & \textbf{68.2} & 59.7 & 52.7 & 59.3 & & 47.3 & 7.0 & 9.0 & 22.0 & 13.7 & \textbf{49.0} & & 59.6 & 23.9 & 57.3 & \textbf{60.9} & 51.8 & 60.0 \\
\quad SlideVQA & 84.2 & 19.2 & \textbf{90.5} & 86.1 & 80.9 & 89.4 & & 76.2 & 5.0 & 4.8 & 66.2 & 30.4 & \textbf{84.3} & & 88.0 & 24.9 & 82.4 & 88.9 & 77.9 & \textbf{91.6} \\
\midrule
Average & 60.5 & 38.6 & 61.6 & 59.6 & 53.1 & \textbf{62.7} & & 35.0 & 10.4 & 12.4 & 27.9 & 20.7 & \textbf{45.4} & & 59.6 & 34.2 & 59.0 & 59.3 & 53.1 & \textbf{62.4} \\
\bottomrule
\end{tabular}
\end{table*}

\section{Detailed results}
\label{sec:suppl_results}
\label{sec:suppl_accuracy}

\paragraph{Text needles.}
On the text needles of the calibration documents (\cref{sec:suppl_calib}), the selector's head at layer 7 puts a needle among the top three text units for 62\% of the questions, against 100\% for the best layer of Qwen3.5-9B (layer 23). A random ranking reaches 0.3\%. For image needles, the selector and Qwen3.5-9B are close (\cref{fig:finding}), so the head works best when the evidence is an image.

\paragraph{Per-dataset accuracy.}
\Cref{tab:datasets} gives the accuracy on each of the 19 datasets.
With \method, it rises most on MM-NIAH counting on every target, and falls most on LongDocURL on Qwen3.5-9B and Qwen3-VL-32B.

\paragraph{Gemma-4-12B and the reasoning limit.}
\Cref{sec:exp_main} makes two statements about the 32K reasoning limit: Gemma-4-12B's full input often reaches it, and \method still raises every target's mean score on the questions where neither the full input nor \method reaches it.
This paragraph gives the numbers behind both.
With the full input, 435 of the 1,750 answers of Gemma-4-12B reach the 32K reasoning limit, and 356 of them are wrong.
The capped answers score 18.4 on average, against 39.3 for the others.
With \method, 266 answers reach the limit.

On the 1,161 questions where neither run reaches it, \method raises the mean score from 42.2 to 55.6.
The gain holds in four categories: from 55.0 to 56.3 in visual RAG, from 48.8 to 80.4 in needle in a haystack, from 1.9 to 27.9 in many-shot ICL (154 questions) and from 62.6 to 63.5 in long-document VQA.
Summarization falls from 22.6 to 18.8.
So many-shot ICL stays low with the full input even when the reasoning ends.

The other two targets also rise: on the questions where neither run reaches the limit, the full input and \method score 67.3 and 70.6 on Qwen3.5-9B (1,637 questions) and 64.5 and 69.0 on Qwen3-VL-32B (1,688 questions).
These are means over questions, not the averages of \cref{tab:main}.

\paragraph{Answers without reasoning.}
Every target in \cref{tab:main} reasons before it answers. In \cref{tab:nothink}, the targets answer the same 1,750 questions without reasoning, under the generation settings of MMLongBench.
Qwen3-VL-32B-Thinking always reasons, so this run uses Qwen3-VL-32B-Instruct, with the reference layer of FlexSelect found again on it.
\method keeps the same units as in \cref{tab:main}, and each baseline keeps as many tokens as \method; FlexSelect and SpecPrefill score the prompt without reasoning again.
Without reasoning, \method stays above the full input on all three targets, by 2.0, 8.1 and 2.6 points, against 2.2, 10.4 and 2.8 points with reasoning.

\begin{table}[t]
\centering
\caption{Average accuracy without and with reasoning, 10\% of the input tokens kept. \emph{Off}: the target answers without reasoning, under the generation settings of MMLongBench (greedy, at most 128 new tokens, 384 for summarization); for Qwen3-VL-32B, this column uses Qwen3-VL-32B-Instruct. \emph{On}: the averages of \cref{tab:main}. \textbf{Bold}: best pruning method.}
\label{tab:nothink}
\footnotesize
\setlength{\tabcolsep}{3.5pt}
\begin{tabular}{@{}l cc cc cc@{}}
\toprule
 & \multicolumn{2}{c}{Qwen3.5-9B} & \multicolumn{2}{c}{Gemma-4-12B} & \multicolumn{2}{c}{Qwen3-VL-32B} \\
\cmidrule(lr){2-3}\cmidrule(lr){4-5}\cmidrule(lr){6-7}
Method & Off & On & Off & On & Off & On \\
\midrule
Full input & 55.4 & 60.5 & 30.5 & 35.0 & 57.0 & 59.6 \\
\midrule
FastV~\cite{chen2024image} & 35.7 & 38.6 & \phantom{0}8.1 & 10.4 & 36.6 & 34.2 \\
FlexSelect~\cite{zhang2025flexselect} & 56.8 & 61.6 & 10.7 & 12.4 & 58.6 & 59.0 \\
SGP~\cite{zhao2025stitch} & 55.9 & 59.6 & 24.6 & 27.9 & 57.5 & 59.3 \\
SpecPrefill~\cite{liu2025speculative} & 55.5 & 53.1 & 18.1 & 20.7 & 55.9 & 53.1 \\
\rowcolor{ourrow} \method (ours) & \textbf{57.4} & \textbf{62.7} & \textbf{38.6} & \textbf{45.4} & \textbf{59.6} & \textbf{62.4} \\
\bottomrule
\end{tabular}
\end{table}

\section{Hardware performance}
\label{sec:suppl_speed}

\paragraph{Time to first token.}
All times are means over the 190 timing questions, each measured on the first request for its input.
A method's time to first token is the sum of \emph{Before} and \emph{Prefill} in \cref{tab:answer}, and its speedup in \cref{tab:main} is the full input's time to first token divided by the method's.
SGLang runs the Qwen targets with FlashAttention-3 and Gemma-4-12B with the Triton backend, the only SGLang backend in which Gemma's image tokens attend to each other in both directions.
For the Qwen targets, we run the image processor before the request and time it as preprocessing; Gemma-4-12B's images go to SGLang as they are, and SGLang processes them while handling the request.
\Cref{tab:answer} therefore counts this step in \emph{Prefill}, or in \emph{Before} for FlexSelect, SGP and SpecPrefill, whose scoring passes run in SGLang too.
FlexSelect's selection is costly: on Qwen3-VL-32B, the image preprocessing and its scoring pass (the vision encoder and the target up to its reference layer 49) take 15.26\,s (\emph{Before} in \cref{tab:answer}), four fifths of the full input's 19.06\,s to the first token, which alone limits its speedup.

\begin{table}[h]
\centering
\caption{Time to answer on the 190 timing questions, in seconds, first request per input. \emph{Before}: steps before the target; \emph{Prefill}: request to first token; \emph{Decode}: first to last token; \emph{Tokens}: mean generated tokens; \emph{Speedup}: full input's Total over the method's.}
\label{tab:answer}
\footnotesize
\setlength{\tabcolsep}{2.6pt}
\begin{tabular}{@{}l rrrr r r r@{}}
\toprule
Method & Before & Prefill & Decode & Total & Speedup & Tokens & ms/tok \\
\midrule
\multicolumn{8}{c}{Qwen3.5-9B} \\
\midrule
Full input & 1.37 & 5.82 & 56.3 & 63.5 & 1.00$\times$ & 6,924 & 8.1 \\
FastV & 1.37 & 3.06 & 67.5 & 71.9 & 0.88$\times$ & 9,152 & 7.4 \\
FlexSelect & 5.38 & 2.25 & 44.4 & 52.0 & 1.22$\times$ & 6,322 & 7.0 \\
SGP & 4.22 & 2.44 & 48.0 & 54.7 & 1.16$\times$ & 6,625 & 7.2 \\
SpecPrefill & 5.03 & 1.88 & 58.0 & 65.0 & 0.98$\times$ & 8,249 & 7.0 \\
\rowcolor{ourrow} \method & 1.45 & 0.48 & 27.3 & 29.2 & 2.17$\times$ & 3,901 & 7.0 \\
\midrule
\multicolumn{8}{c}{Gemma-4-12B} \\
\midrule
Full input & -- & 34.05 & 366.2 & 400.2 & 1.00$\times$ & 11,115 & 32.9 \\
FastV & -- & 9.83 & 294.7 & 304.5 & 1.31$\times$ & 17,618 & 16.7 \\
FlexSelect & 8.79 & 5.31 & 313.3 & 327.4 & 1.22$\times$ & 21,155 & 14.8 \\
SGP & 26.18 & 6.47 & 214.1 & 246.8 & 1.62$\times$ & 13,669 & 15.7 \\
SpecPrefill & 30.58 & 4.88 & 344.3 & 379.7 & 1.05$\times$ & 23,140 & 14.9 \\
\rowcolor{ourrow} \method & 1.35 & 1.59 & 103.4 & 106.4 & 3.76$\times$ & 7,003 & 14.8 \\
\midrule
\multicolumn{8}{c}{Qwen3-VL-32B} \\
\midrule
Full input & 1.48 & 17.58 & 35.7 & 54.7 & 1.00$\times$ & 1,898 & 18.8 \\
FastV & 1.48 & 3.23 & 83.3 & 88.0 & 0.62$\times$ & 5,524 & 15.1 \\
FlexSelect & 15.26 & 2.42 & 74.6 & 92.3 & 0.59$\times$ & 5,059 & 14.7 \\
SGP & 6.00 & 3.08 & 65.1 & 74.2 & 0.74$\times$ & 4,296 & 15.2 \\
SpecPrefill & 6.20 & 2.11 & 58.4 & 66.7 & 0.82$\times$ & 3,973 & 14.7 \\
\rowcolor{ourrow} \method & 1.46 & 0.87 & 20.6 & 22.9 & 2.39$\times$ & 1,443 & 14.3 \\
\bottomrule
\end{tabular}
\end{table}

\paragraph{Time to answer.}
For \cref{tab:answer}, the target of each method answers the 190 timing questions again and generates exactly as many tokens as its reasoning and answer in the accuracy run of \cref{tab:main}, so that its decoding takes as long as in that run.
We decode greedily in one streamed request and do not stop at the end-of-sequence token until this count is reached.
The steps before the target are timed in separate runs on the same questions.
Since the decoding takes most of the time, \method gains from generating fewer tokens and from its shorter KV cache.

\paragraph{Selector on the target's GPUs.}

The selector can also share the target's GPUs (\cref{tab:colocated}).
We run it and the target as two SGLang servers on the same H100 (on Qwen3-VL-32B, the same two GPUs, with the selector on the first), and they take turns on each question.
The selector gets 6\% of the GPU memory, taken from the target's usual amount.
Sharing changes the selector's time by at most 0.02\,s and the target's by less than 0.01\,s, so the speedups change by at most 0.1$\times$.

\begin{table}[h]
\centering
\caption{The selector on its own H100 (\emph{Separate}) or on the target's GPUs (\emph{Shared}), 190 timing questions. Times in seconds; \emph{Before} and \emph{Prefill} as in \cref{tab:answer}, \emph{Total}: time to first token (Before plus Prefill), \emph{Speedup}: as in \cref{tab:main}. The \emph{Separate} rows take Before and Prefill from \cref{tab:answer}; the \emph{Shared} rows add the change measured with the selector and the target on the same GPUs in an earlier run on the same questions.}
\label{tab:colocated}
\footnotesize
\setlength{\tabcolsep}{3pt}
\begin{tabular}{@{}ll rrr r@{}}
\toprule
Target & GPUs & Before & Prefill & Total & Speedup \\
\midrule
Qwen3.5-9B & Separate & 1.45 & 0.48 & 1.93 & 3.7$\times$ \\
 & Shared & 1.46 & 0.48 & 1.94 & 3.7$\times$ \\
\addlinespace[2pt]
Gemma-4-12B & Separate & 1.35 & 1.59 & 2.95 & 11.6$\times$ \\
 & Shared & 1.35 & 1.59 & 2.94 & 11.6$\times$ \\
\addlinespace[2pt]
Qwen3-VL-32B & Separate & 1.46 & 0.87 & 2.33 & 8.2$\times$ \\
 & Shared & 1.48 & 0.87 & 2.35 & 8.1$\times$ \\
\bottomrule
\end{tabular}
\end{table}

\section{More ablations}
\label{sec:suppl_ablation}

\paragraph{Selector resolution.}
\Cref{tab:resolution} gives the accuracy and speedup of every target with the selector reading the images at full and at half resolution.
\begin{table}[t]
\centering
\caption{Selector images at full or half resolution, 10\% of the input tokens kept; the target reads the kept images at full resolution.}
\label{tab:resolution}
\footnotesize
\setlength{\tabcolsep}{3pt}
\begin{tabular}{@{}l c cc cc@{}}
\toprule
& Full input & \multicolumn{2}{c}{Selector, full res.} & \multicolumn{2}{c}{Selector, half res.\ (ours)} \\
\cmidrule(lr){3-4}\cmidrule(l){5-6}
Target & Avg. & Avg. & Speedup & Avg. & Speedup \\
\midrule
Qwen3.5-9B & 60.5 & 63.4 & 2.0$\times$ & 62.7 & 3.7$\times$ \\
Gemma-4-12B & 35.0 & 45.5 & 7.1$\times$ & 45.4 & 11.6$\times$ \\
Qwen3-VL-32B & 59.6 & 62.2 & 4.8$\times$ & 62.4 & 8.2$\times$ \\
\bottomrule
\end{tabular}
\end{table}

\paragraph{Larger selectors.}
In \cref{tab:larger}, Qwen3.5-2B or Qwen3-VL-2B, a full-attention model, replaces the 0.8B selector, each calibrated on the same 50 documents.
They raise the average by up to 0.6 points, but they run 12 layers instead of 8 and lower the speedup on both targets; at the same depth, the full-attention Qwen3-VL-2B selects more slowly than the hybrid Qwen3.5-2B.
\begin{table}[t]
\centering
\caption{Larger selectors, 10\% of the input tokens kept. \emph{Avg.}\ and \emph{Speedup} as in \cref{tab:main}; \emph{Select}: the selector's time per question on Qwen3.5-9B, in seconds, preprocessing excluded.}
\label{tab:larger}
\footnotesize
\setlength{\tabcolsep}{3pt}
\begin{tabular}{@{}l c rr rr r@{}}
\toprule
 & & \multicolumn{2}{c}{Qwen3.5-9B} & \multicolumn{2}{c}{Gemma-4-12B} & \\
\cmidrule(lr){3-4}\cmidrule(lr){5-6}
Selector & Layers & Avg. & Speedup & Avg. & Speedup & Select \\
\midrule
Full input & -- & 60.5 & 1.0$\times$ & 35.0 & 1.0$\times$ & -- \\
\rowcolor{ourrow} Qwen3.5-0.8B (ours) & 8 & 62.7 & 3.7$\times$ & 45.4 & 11.6$\times$ & 0.71 \\
Qwen3.5-2B & 12 & 62.7 & 3.2$\times$ & 45.5 & 10.3$\times$ & 0.96 \\
Qwen3-VL-2B & 12 & 62.7 & 2.9$\times$ & 46.0 & 9.6$\times$ & 1.20 \\
\bottomrule
\end{tabular}
\end{table}

\paragraph{Retrievers as scorers.}

\Cref{tab:retrieval} lists the accuracy of each scorer per category, with its speedup.
The retrievers fall behind the head most in VRAG, by 3.0 to 9.0 points, while ColEmbed V2 8B is ahead in NIAH, Summ and DocQA.

\Cref{tab:retrieval_evidence} measures, as \cref{fig:evidence} does, how much of the annotated evidence each scorer keeps.
\begin{table}[t]
\centering
\caption{Whole units ranked by each scorer (paragraphs by BM25~\cite{robertson2009bm25} on the question text, except for GME-2B and \method), Qwen3.5-9B, 10\% kept; \textbf{bold}: best.}
\label{tab:retrieval}
\footnotesize
\setlength{\tabcolsep}{2.1pt}
\begin{tabular}{l*{6}{r}r}
\toprule
Scorer & VRAG & NIAH & ICL & Summ & DocQA & Avg. & Speedup \\
\midrule
Full input & 62.0 & 60.5 & 85.5 & 22.5 & 72.1 & 60.5 & 1.0$\times$ \\
\midrule
SigLIP 2 & 55.0 & 72.6 & \textbf{91.3} & 18.0 & 49.8 & 57.3 & \textbf{4.0$\times$} \\
ColEmbed V2 8B & 57.5 & \textbf{77.0} & 86.8 & \textbf{19.8} & \textbf{70.9} & 62.4 & 1.3$\times$ \\
GME-2B & 61.0 & 65.6 & 90.0 & 18.8 & 70.5 & 61.2 & 1.6$\times$ \\
\rowcolor{ourrow} \method (ours) & \textbf{64.0} & 75.0 & 87.0 & 18.3 & 69.4 & \textbf{62.7} & 3.7$\times$ \\
\bottomrule
\end{tabular}
\end{table}

\begin{table}[t]
\centering
\caption{Annotated evidence tokens kept at a 10\% budget (\%), on the questions of \cref{fig:evidence} (VH: Visual Haystacks); BM25 ranks the paragraphs for all scorers but GME-2B and \method. \textbf{Bold}: best per column.}
\label{tab:retrieval_evidence}
\footnotesize
\setlength{\tabcolsep}{2.25pt}
\begin{tabular}{l*{4}{r}}
\toprule
Scorer & NIAH, image & NIAH, text & VH & DocQA \\
\midrule
SigLIP 2 & 96.0 & 88.4 & 41.3 & 54.3 \\
ColEmbed V2 8B & \textbf{96.7} & 88.4 & 67.8 & \textbf{85.7} \\
GME-2B & 94.0 & 78.4 & 41.8 & 78.3 \\
\rowcolor{ourrow} \method (ours) & \textbf{96.7} & \textbf{92.0} & \textbf{71.5} & 79.4 \\
\bottomrule
\end{tabular}
\end{table}

SigLIP 2 keeps the needle images about as well as \method, but it loses most Visual Haystacks needles (41.3 against 71.5) and many answer pages of long documents (54.3 against 79.4).
ColEmbed V2, trained on document pages, keeps the most answer pages (85.7) but fewer Visual Haystacks needles (67.8).
For text, BM25 keeps fewer needles than the head (88.4 against 92.0).

\section{Failure analysis}
\label{sec:suppl_failure}

\paragraph{LongDocURL.}
This analysis splits the 100 LongDocURL questions of \cref{tab:main} (10\% of the input tokens kept) by how many of their annotated answer pages \method keeps: every one for 54 questions, some for 28 and none for 18 (the \emph{Missed} column of \cref{tab:longdocurl}).
On LongDocURL, \method scores 59.3 against SGP's 73.8 on Qwen3.5-9B and 62.5 against 73.9 on Qwen3-VL-32B (\cref{tab:datasets}), and almost all of this gap comes from the 18 questions.
On the 54 questions, both methods score about the same: 78.7 and 80.3 for \method on Qwen3.5-9B and Qwen3-VL-32B, against 79.9 and 81.0 for SGP.
On the 18 questions, \method scores 3.6 and 0.9, and SGP 64.7 and 59.6: SGP keeps a few tokens on almost every page, and so some tokens of an answer page for all 100 questions.
If \method matched SGP on these 18 questions, its gap would shrink from 14.5 to 3.5 points on Qwen3.5-9B and from 11.4 to 0.8 on Qwen3-VL-32B.

\paragraph{FlexSelect's kept tokens on Gemma-4-12B.}
On Gemma-4-12B, FlexSelect's kept tokens spread evenly over the document (\cref{sec:exp_evidence}).
To measure this, we cut each document into ten parts of equal length, from its start to its end, and count how many of the kept tokens fall into each part.
On Gemma-4-12B, every part holds 9.4 to 10.3\% of the kept tokens.
On the other two targets, FlexSelect favors one end: on Qwen3.5-9B, the first part holds 19.4\% of the kept tokens, and on Qwen3-VL-32B the last part holds 19.4\%.

\paragraph{SGP's small models.}
SGP's low accuracy on Gemma-4-12B comes mostly from keeping single tokens (\cref{sec:exp_failure}); its small model, Gemma-4-E2B, is not the cause.
Outside summarization, which the small models' 64-token answers cut short, Gemma-4-E2B answers fewer questions correctly than Qwen3.5-2B (620 against 834 of the 1,550 questions).
Yet even on these 620 questions, Gemma-4-12B averages 35.3 with SGP, below both the full input (46.7) and \method (66.6).

\paragraph{Many-shot ICL on Gemma-4-12B.}
\begin{table}[t]
\centering
\caption{Accuracy per category with the full input and with \method at budgets of 5, 10 and 15\% (rows 5--15\%; selector at half resolution), on the 1,750 test questions. \emph{Tokens}: mean generated tokens in many-shot ICL; \emph{32K}: many-shot ICL answers, of 400, that reach the 32K reasoning limit.}
\label{tab:budget_categories}
\footnotesize
\setlength{\tabcolsep}{2.3pt}
\begin{tabular}{l*{6}{r}rr}
\toprule
 & \multicolumn{6}{c}{Accuracy} & \multicolumn{2}{c}{Many-shot ICL} \\
\cmidrule(lr){2-7}\cmidrule(l){8-9}
Input & VRAG & NIAH & ICL & Summ & DocQA & Avg. & Tokens & 32K \\
\midrule
\multicolumn{9}{l}{\emph{Qwen3-VL-32B}} \\
Full input & 60.0 & 58.5 & 82.8 & 22.3 & 74.6 & 59.6 & 0.7K & 5 \\
5\% & 62.5 & 71.7 & 81.3 & 17.3 & 59.6 & 58.5 & 1.0K & 2 \\
\rowcolor{ourrow} 10\% & 67.5 & 69.4 & 84.8 & 18.9 & 71.4 & 62.4 & 1.3K & 4 \\
15\% & 64.0 & 71.2 & 85.8 & 19.8 & 71.9 & 62.5 & 1.1K & 5 \\
\midrule
\multicolumn{9}{l}{\emph{Gemma-4-12B}} \\
Full input & 50.0 & 39.1 & 2.8 & 22.6 & 60.4 & 35.0 & 14.9K & 157 \\
5\% & 50.5 & 66.7 & 46.0 & 17.0 & 55.9 & 47.2 & 5.0K & 32 \\
\rowcolor{ourrow} 10\% & 54.5 & 65.2 & 26.3 & 18.8 & 62.1 & 45.4 & 14.7K & 149 \\
15\% & 54.0 & 60.8 & 18.3 & 19.9 & 64.0 & 43.4 & 17.3K & 184 \\
\bottomrule
\end{tabular}
\end{table}

\Cref{sec:exp_budget} notes that Gemma-4-12B's many-shot ICL score falls as the budget grows, and \cref{sec:exp_failure} that token-level selection splits the examples apart; the three observations below give the details.

First, Gemma-4-12B scores higher on many-shot ICL the less of the input it reads (\cref{tab:budget_categories}): 46.0 with 5\% of the tokens, 26.3 with 10\%, 18.3 with 15\% and 2.8 with the full input.
With more input, its answers grow longer (from 5.0K to 17.3K tokens between 5\% and 15\%) and more of them reach the 32K limit (from 32 to 184 of 400; 157 with the full input).
The limit does not explain the lower scores, since at every budget the answers that reach it score no worse than the others (59.4 against 44.8 at 5\%, 19.0 against 17.6 at 15\%).

Second, token-level selection breaks the examples apart.
Each example is an image followed by a label line (\texttt{label: N}).
FlexSelect and SGP keep a few tokens of nearly every example image (8.7\% and 7.5\% of its tokens on average) but few label lines whole (almost none and 35\%), whereas \method keeps 9.8\% of the examples, each image whole together with its label.
A label line counts as kept whole when all its tokens are kept; both measures are averaged within each question and then over the 400 questions.

\begin{figure}[t]
\centering
\includegraphics[width=\linewidth]{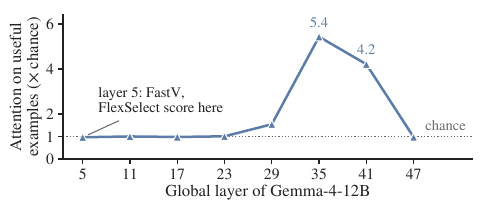}
\caption{Attention of Gemma-4-12B's last prompt token to the useful many-shot ICL examples, those whose label is the answer, at each global layer, averaged over the heads and relative to an even spread of the attention over all examples (1: chance); full input, 50 questions.}
\label{fig:icl_attention}
\end{figure}
Third, FastV and FlexSelect score Gemma-4-12B at layer 5, which may not suit many-shot ICL (\cref{sec:suppl_baselines}).
FastV takes it as the first layer that attends to the whole input, and FlexSelect picks it with a needle test on MM-NIAH, which rewards finding an inserted image that matches the question.
Many-shot ICL labels its examples with numbers, so the target can find the answer only by matching the image to classify with the examples of its class and reading their label; the needle test does not check whether a layer finds these examples.
On 50 of the 400 questions, we measure how much attention the last prompt token gives these examples at each global layer, relative to an even spread over all examples (\cref{fig:icl_attention}).
The ratio stays at chance up to layer 23 and reaches 5.4 and 4.2 only at layers 35 and 41: the target finds the useful examples in its late layers, which FastV and FlexSelect could use only by running most of the target first.

\end{document}